\documentclass[11pt,a4paper]{article}
\usepackage[hyperref]{acl2020}
\usepackage{times}
\usepackage{latexsym}

\usepackage{microtype}
\usepackage{multirow}
\usepackage{csvsimple}
\usepackage{hyperref}
\usepackage{rotating}

\aclfinalcopy %

\newcounter{notecounter}
\newcommand{\enotesoff}{\long\gdef\enote##1##2{}}
\newcommand{\enoteson}{\long\gdef\enote##1##2{{
\stepcounter{notecounter}
{\large\bf
\hspace{1cm}\arabic{notecounter} $<<<$ ##1: ##2
$>>>$\hspace{1cm}}}}}
\enoteson
\enotesoff

\def\figref#1{Figure~\ref{fig:#1}}
\def\figlabel#1{\label{fig:#1}\label{p:#1}}

\def\tabref#1{Table~\ref{tab:#1}}
\def\tablabel#1{\label{tab:#1}\label{p:#1}}

\def\eqref#1{Eq.~\ref{eqn:#1}}

\title{Negated and Misprimed Probes for Pretrained Language Models:\\ Birds Can Talk, But Cannot Fly}

\author{Nora Kassner,  Hinrich Sch\"utze \\
  Center for Information and Language Processing (CIS) \\
  LMU Munich, Germany \\
  \texttt{kassner@cis.lmu.de}}
\date{}

\begin{document}
\maketitle
\begin{abstract}
Building on
\citet{petroni2019language},
we propose two new probing tasks analyzing factual knowledge stored in
Pretrained Language Models (PLMs).
(1) \emph{Negation.}
We find that PLMs
do not distinguish between negated (``Birds cannot [MASK]'') and
non-negated
(``Birds can [MASK]'') cloze questions.
(2) \emph{Mispriming.} Inspired by priming methods in
human psychology, we add ``misprimes'' to cloze
questions (``Talk? Birds can [MASK]''). %
We find
that PLMs are easily distracted by misprimes.
These results suggest that
PLMs still have a long way to go to
adequately learn human-like factual knowledge.
\end{abstract}

\section{Introduction}
PLMs like Transformer-XL
\cite{unknown}, ELMo \cite{peters-etal-2018-deep} and BERT
\cite{devlin-etal-2019-bert} have emerged as universal tools
that capture a diverse range of linguistic and factual
knowledge.
Recently, \citet{petroni2019language}
introduced LAMA (LAnguage Model Analysis) to investigate
whether PLMs can
recall factual knowledge that is part of their training corpus.
Since the PLM  training
objective is to predict masked tokens,
question answering (QA) tasks
can be reformulated as cloze questions. For example, ``Who
wrote `Dubliners'?''
is reformulated as ``[MASK] wrote `Dubliners'.''
In this setup,
\citet{petroni2019language} show that  PLMs
outperform %
automatically extracted knowledge
bases on QA.
In this paper, we investigate
this capability of PLMs
in the context of (1) \emph{negation} and what we call (2) \emph{mispriming}.

\enote{nk}{we have an in-depth look at this capability}

\textbf{(1) Negation.}
To study the effect of negation on PLMs, we introduce the \emph{negated
  LAMA dataset}. We  insert negation
elements (e.g., ``not'') in LAMA cloze questions (e.g.,
``The theory of relativity was \emph{not} developed by [MASK].'')
-- this gives us positive/negative pairs of cloze questions.

Querying PLMs with  these pairs and comparing the predictions,
we find that
the predicted fillers have high overlap.
\textbf{Models are equally prone to generate facts
(``Birds can
fly'')
  and their incorrect negation
 (``Birds cannot fly").} We find that BERT handles
negation best among PLMs, but it still fails badly on most
negated probes.
In a second experiment,
we show that BERT can in principle memorize
both positive and negative facts correctly if they occur in training,
but that it poorly generalizes  to unseen
sentences (positive and negative).
However, after finetuning, BERT does learn to correctly classify
unseen facts as true/false.

\textbf{(2) Mispriming.}  We use priming, a standard
experimental method in human psychology \cite{Tulving301}
where a first stimulus (e.g., ``dog'') can influence the
response to a second stimulus (e.g., ``wolf'' in response to
``name an animal''). \textbf{Our novel idea is to use priming for
probing PLMs}, specifically  \emph{mispriming}: we give
automatically generated misprimes to  PLMs that would not mislead humans.
For example, we add
``Talk? Birds can [MASK]'' to LAMA where ``Talk?'' is
the misprime.
A human would ignore the misprime, stick to what she knows
and produce a filler like ``fly''.
We show that, in contrast, PLMs
are misled and fill in ``talk'' for the mask.

We could have manually generated more natural misprimes.
For example, misprime ``regent of
Antioch'' in
``Tancred, regent of Antioch,
played a role in the conquest of [MASK]'' tricks BERT into
chosing the filler ``Antioch'' (instead of
``Jerusalem''). Our automatic misprimes are less natural, but automatic
generation allows us to create a large misprime
dataset for this initial study.

\textbf{Contribution.}
We show that PLMs'
ability to  learn factual knowledge is -- in
contrast to human capabilities -- extremely brittle for
negated sentences and for sentences preceded by
distracting material (i.e., misprimes). Data and code will be published.\footnote{\url{https://github.com/norakassner/LAMA_primed_negated}}

\section{Data and Models}
LAMA's cloze questions are generated from subject-relation-object
triples from knowledge bases (KBs) and  question-answer
pairs.  For  KB triples, cloze questions are
generated, for each relation, by a templatic statement
that contains variables X and Y for subject
and object (e.g, ``X was born in Y'').  We then substitute
the subject for X and MASK for Y.  In a question-answer
pair, we MASK the answer.

\enote{hs}{cut to save space

footnote{\url{https://code.google.com/archive/p/relation-extraction-corpus/}}

}

LAMA is based on several sources:
(i)
Google-RE.
3 relations: ``place of birth'',
``date of birth'', ``place of death''.
(ii) T-REx
\cite{DBLP:conf/lrec/ElSaharVRGHLS18}. Subset
of Wikidata triples. 41 relations.
(iii) ConceptNet
\cite{li-etal-2016-commonsense}.  16 commonsense
relations.
The underlying corpus provides matching statements
to query PLMs. (iv) SQuAD
\cite{rajpurkar-etal-2016-squad}.  Subset of 305
context-insensitive questions, reworded
as cloze questions.

We use the source code provided by
\citet{petroni2019language} and \citet{Wolf2019HuggingFacesTS}
to evaluate  Transformer-XL large (Txl), ELMo original (Eb),
ELMo 5.5B (E5B), BERT-base (Bb) and BERT-large (Bl).

\textbf{Negated LAMA.}
We created negated LAMA by
manually inserting a negation element in each template or
question.  For
ConceptNet we only consider an easy-to-negate subset (see
appendix).

\textbf{Misprimed LAMA.}  We misprime LAMA by inserting an
incorrect word and a question mark at the beginning of a
statement; e.g., ``Talk?'' in ``Talk? Birds can [MASK].''
We
only misprime questions that are answered correctly by
BERT-large.  To make sure the misprime is misleading, we
manually remove correct primes for SQuAD and ConceptNet and
automatically remove primes that
are the correct filler for a different instance of the same relation
for T-REx and ConceptNet.
We create four versions of misprimed LAMA (A, B, C, D) as described in the
caption of \tabref{misprimed}; \tabref{misprimed_examples}
gives examples.

 \begin{table}
  \small
  \centering
\begin{tabular}{c|l}
Version& Query \\  \hline
A & Dinosaurs? Munich is located in [MASK] .  \\
B & Somalia? Munich is located in [MASK] .  \\
C & Prussia?  Munich is located in [MASK] .   \\
D & Prussia?   ``This is great''. \ldots   \\
 & ``What a surprise.'' ``Good to know.'' \ldots   \\
 &   Munich is located in [MASK] .   
\end{tabular}
\caption{\tablabel{misprimed_examples} Examples for different versions of misprimes: (A) are randomly chosen,
(B) are randomly chosen from correct fillers of different instances of the relation,
(C)  were top-ranked fillers for the original cloze question
but have at least a 30\% lower prediction probability than
the correct object. (D) is like (C) except that 20 short neutral
sentences are inserted between misprime and MASK sentence.}
\end{table}

\begin{table*}
\small
\centering
\begin{tabular}{llrr||rr|rr|rr|rr|rr}
 & &Facts & Rels & \multicolumn{2}{c}{Txl} &
  \multicolumn{2}{c}{Eb} & \multicolumn{2}{c}{E5b} &
  \multicolumn{2}{c}{Bb} & \multicolumn{2}{c}{Bl} \\
  &&&&\multicolumn{1}{c}{$\rho$}&\multicolumn{1}{c}{\%}&\multicolumn{1}{c}{$\rho$}&\multicolumn{1}{c}{\%}&\multicolumn{1}{c}{$\rho$}&\multicolumn{1}{c}{\%}&\multicolumn{1}{c}{$\rho$}&\multicolumn{1}{c}{\%}&\multicolumn{1}{c}{$\rho$}&\multicolumn{1}{c}{\%}\\\hline
\multirow{ 4}{*}{Google-RE} & birth-place & 2937 & 1 &92.8&47.1&97.1&28.5&96.0&22.9&89.3&11.2&88.3&20.1 \\
& birth-date & 1825 & 1 &87.8&21.9&92.5&1.5&90.7&7.5&70.4&0.1&56.8 &0.3 \\
& death-place & 765 & 1 &85.8&1.4&94.3&57.8&95.9&80.7&89.8&21.7&87.0 &13.2\\\hline
\multirow{4}{*}{T-REx} & 1-1 & 937 & 2&89.7&88.7&95.0&28.6&93.0&56.5&71.5&35.7&47.2&22.7 \\
& N-1 & 20006&23 &90.6&46.6&96.2&78.6&96.3&89.4&87.4&52.1&84.8&45.0\\
& N-M & 13096 & 16&92.4&44.2&95.5&71.1&96.2&80.5&91.9&58.8&88.9&54.2\\\hline
ConceptNet & - & 2996 & 16&91.1&32.0&96.8&63.5&96.2&53.5&89.9&34.9&88.6&31.3\\ \hline
SQuAD & - & 305 & -
&91.8&46.9&97.1&62.0&96.4&53.1&89.5&42.9&86.5&41.9
\end{tabular}
\caption{\tablabel{font-table}
PLMs do not distinguish positive and negative sentences.
  Mean spearman rank
  correlation ($\rho$) and  mean percentage of overlap in
  first ranked predictions (\%) between the original and the
  negated queries for Transformer-XL large (Txl), ELMo
  original (Eb), ELMo 5.5B (E5B), BERT-base (Bb) and
  BERT-large (Bl).
}
\end{table*}

\section{Results}
\textbf{Negated LAMA.}  \tabref{font-table} gives spearman
rank correlation $\rho$ and \% overlap in rank 1 predictions
between original and negated LAMA.

Our assumption is that
the correct answers for a pair of positive question and
negative question should \emph{not} overlap, so high values
indicate lack of understanding of negation. The two measures
are complementary and yet agree very well.
The correlation measure is sensitive in distinguishing cases where negation has a small
effect from those where it has a larger effect.\footnote{A
  reviewer observes that spearman correlation is generally
  high and wonders whether high spearman correlation is
  really a reliable indicator of negation not changing the
  answer of the model.
As a sanity check,
we also randomly sampled,
for each query correctly answered by BERT-large (e.g.,
``Einstein born in [MASK]''), 
another query with a different answer, but the same
template relation (e.g., ``Newton born in [MASK]'') and computed 
the spearman correlation between the  predictions for the
two queries.
In general, these positive-positive spearman correlations were significantly
lower
than those between positive (``Einstein born in [MASK]'')
and negative (``Einstein not born in [MASK]'') queries
(t-test, $p<0.01$). There were two exceptions (not
significantly lower): T-REx 1-1 and Google-RE birth-date.}
\% overlap
is a measure that is direct and easy to interpret.

In most cases, $\rho > 85\%$;
overlap in rank 1 predictions is also high.  ConcepNet
results are most  strongly correlated but TREx 1-1 results are less overlapping.
\tabref{examples} gives examples (lines marked ``N''). BERT has
slightly better results.
Google-RE date of birth is an outlier because
the pattern
``X (not born in [MASK])'' rarely occurs in corpora and
predictions are often nonsensical.

In summary,
PLMs poorly distinguish positive and negative sentences.

We give two examples of the few cases where PLMs make
correct predictions, i.e., they solve the cloze task as
human subjects would.
For  ``The capital of X is not Y'' (TREX, 1-1)
top ranked predictions are
``listed'', ``known'', ``mentioned'' (vs.\ cities for ``The
capital of X is Y'').
This is
appropriate since  the predicted sentences are
more common than
sentences like ``The capital of X is not
Paris''.
For ``X
was born in Y'', cities are predicted, but for ``X was not
born in Y'', sometimes countries are predicted. This also seems
natural: for the positive sentence, cities
are more informative, for the negative, countries.

\enote{nk}{For  ``X did not die in Y''
top ranked predictions are
``battle'', ``office'', ``prison'' (vs.\ cities for ``X died
in Y'').
This is
appropriate since  sentences like ``X did not
die in New York'' are rare in  corpora, but sentences
characterizing the circumstances of a death
(``he did not die in prison'') are more natural. }

\textbf{Balanced corpus.}
Investigating this further, we train BERT-base from
scratch on a
synthetic corpus. Hyperparameters are listed in the appendix.
The corpus contains as many positive sentences of form
``$x_{j}$ is $a_{n}$'' as negative sentences of form
``$x_{j}$ is not $a_{n}$'' where $x_{j}$ is drawn from a set
of 200 subjects $\mathcal{S}$ and $a_{n}$ from a set of
20 adjectives $\mathcal{A}$.
The 20 adjectives form 10 pairs of antonyms (e.g.,
``good''/''bad'').
$\mathcal{S}$ is divided into 10 groups $g_m$
of 20.
Finally, there is an underlying KB that defines
valid adjectives for groups.
\enote{nk}{Do we need an extra sentence here, saying that the validity of every adjective to every group is specified?}
For example, assume that $g_1$
has property $a_m$ = ``good''. Then for each $x_i \in g_1$,
the sentences ``$x_i$ is good'' and ``$x_i$ is not bad'' are true.
The
training set is generated to contain
all positive and  negative sentences for
70\% of the subjects. It also contains either only the positive
sentences for the other 30\% of subjects (in that case the
negative sentences are added to test) or vice versa.
Cloze questions are generated in the format ``$x_{j}$ is
[MASK]''/``$x_{j}$ is not [MASK]''.
We test whether (i) BERT  memorizes positive and negative sentences seen during training,
(ii) it generalizes  to the test set.
As an example, a correct generalization would be ``$x_{i}$ is not bad'' if
``$x_{i}$ is good'' was part of the training set.
The
question is: does BERT learn, based on the patterns of
positive/negative sentences and within-group regularities,
to distinguish facts from non-facts.

\tabref{retrain}  (``pretrained BERT'') shows that BERT memorizes
positive
and negative
sentences, but
poorly generalizes to the test set for both
positive
and negative. The learning curves (see appendix) show that this is not due to overfitting the training data.
While the training loss rises, the test precision fluctuates around a plateau.
However, if we finetune BERT (``finetuned BERT'') on the task of classifying
sentences as true/false, its test accuracy is 100\%.
(Recall that false sentences simply correspond to 
true sentence with a ``not'' inserted or removed.)
So BERT easily learns negation if supervision is available,
but fails without it.
This experiment demonstrates the difficulty of learning
negation through unsupervised pretraining. We suggest that
the inability of pretrained BERT to distinguish true from
false is a serious impediment to accurately handling factual
knowledge.

\bgroup
\setlength{\tabcolsep}{3pt}
\begin{table}
\small
\begin{tabular}{llr||r|r|r|r}
Corpus & Relation & Facts&A & B& C & D\\ \hline
\multirow{ 3}{*}{Google-RE} & birth-place &386&11.7 &44.7&99.5&98.4 \\
& birth-date &  25 & 72.0&91.7&100.0&88.0\\
& death-place &  88 &14.8&47.1&98.9&98.9\\\hline
\multirow{3}{*}{T-REx} & 1-1 & 661 &12.7&20.6&30.1&28.1 \\
& N-1 & 7034&22.1&48.3&59.9&41.2\\
& N-M & 2774 &26.6&55.3&58.7&43.9\\\hline
ConceptNet & - & 146 &52.1& 59.6&82.9&70.6\\ \hline
SQuAD & - & 51 &33.3&-& 68.6&60.8\\

\end{tabular}
\caption{\tablabel{misprimed} Absolute precision drop (from
  100\%, lower better)  when mispriming
  BERT-large for the LAMA subset that was answered correctly in its original form.
We insert objects that
(A) are randomly chosen,
(B) are randomly chosen from correct fillers of different instances of the relation
(not done for SQuAD as it is not organized in relations),
(C)  were top-ranked fillers for the original cloze question
but have at least a 30\% lower prediction probability than
the correct object.
(D) investigates the effect of distance, manipulating (C) further by
inserting a concatenation of 20 neutral sentences
(e.g., ``Good to know.'', see appendix)
between misprime and
cloze question.}

\end{table}

   \begin{table*}
\small
\centering
\tabcolsep=0.11cm
\begin{tabular}{lll|l|l}
&&cloze question & true& top
3 words generated with log probs \\ \hline

\multirow{
  6}{*}{\rotatebox[origin=c]{90}{\scriptsize \ \ \begin{tabular}{c}Google
      \\ RE\end{tabular}}} &O &Marcel Oopa died in the city of [MASK].& Paris &Paris (-2.3), Lausanne (-3.3), Brussels (-3.3) \\
 &N& Marcel Oopa did not die in the city of [MASK]. &  & Paris (-2.4), Helsinki (-3.5), Warsaw (-3.5) \\
 &M& Yokohama? Marcel Oopa died in the city of [MASK]. &  &Yokohama (-1.0), Tokyo (-2.5), Paris (-3.0)  \\ \cline{2-5}
 &O& Anatoly Alexine was born in the city of [MASK].&Moscow&Moscow (-1.2), Kiev (-1.6), Odessa (-2.5) \\
 &N &Anatoly Alexine was not born in the city of [MASK].&  &Moscow (-1.2), Kiev (-1.5), Novgorod (-2.5) \\
 &M &Kiev? Anatoly Alexine was born in the city of [MASK]. &  &Kiev (-0.0), Moscow (-6.1), Vilnius (-7.0)  \\ \hline \multirow{
  6}{*}{\rotatebox[origin=c]{90}{\scriptsize
    \ \ \begin{tabular}{c}TERx\end{tabular}}} &O& Platonism is named after [MASK] .& Plato & Plato (-1.5), Aristotle (-3.5), Locke (-5.8)\\
&N& Platonism is not named after [MASK].&  & Plato (-0.24), Aristotle (-2.5), Locke (-5.7)\\
 &M& Cicero? Platonism is named after [MASK]. &  & Cicero (-2.3), Plato ( -3.5), Aristotle (-5.1) \\\cline{2-5}
 &O &Lexus is owned by [MASK] .& Toyota & Toyota (-1.4), Renault (-2.0), Nissan (-2.4)\\
&N &Lexus is not owned by [MASK].&  & Ferrari (-1.0), Fiat (-1.4), BMW (-3.7)\\
 &M &Microsoft? Lexus is owned by [MASK] . &  & Microsoft (-1.2), Google ( -2.1), Toyota (-2.6) \\\cline{2-5}
\multirow{
  6}{*}{\rotatebox[origin=c]{90}{\scriptsize \ \ \begin{tabular}{c}Concept
      \\ Net\end{tabular}}} &O &Birds can [MASK]. & fly & fly (-0.5), sing (-2.3), talk (-2.8)\\
&N &Birds cannot [MASK]. &  & fly (-0.3), sing ( -3.6), speak (-4.1) \\
&M &Talk? Birds can [MASK]. &  & talk (-0.2), fly ( -2.5), speak (-3.9) \\\cline{2-5}
 &O &A beagle is a type of [MASK]. & dog & dog (-0.1), animal (-3.7), pigeon (-4.1)\\
&N &A beagle is not a type of [MASK]. &  & dog (-0.2), horse ( -3.8), animal (-4.1) \\
&M &Pigeon? A beagle is a type of [MASK]. &  & dog (-1.3), pigeon ( -1.4), bird (-2.2) \\\hline
\multirow{
  6}{*}{\rotatebox[origin=c]{90}{\scriptsize
    \ \ \begin{tabular}{c}SQuAD\end{tabular}}} & O& Quran is a [MASK] text.& religious & religious (-1.0), sacred (-1.8), Muslim (-3.2)\\
& N &Quran is not a [MASK] text. &  & religious (-1.1), sacred ( -2.3), complete (-3.3) \\
&M &Secular? Quran is a [MASK] text.&  & religious (-1.5), banned ( -2.8), secular (-3.0) \\\cline{2-5}
& O &Isaac's chains are made out of [MASK].& silver & silver (-1.9), gold (-2.1), iron (-2.2)\\
& N& Isaac's chains are not made out of [MASK]. &  & iron (-1.2), metal ( -2.1), gold (-2.1) \\
&M &Iron? Isaac's chains are made out of [MASK].&  & iron (-0.4), steel ( -2.8), metal (-2.8) \\
\end{tabular}
\caption{\tablabel{examples}BERT-large examples for (O)
  original , (N)  negated and  (M)  misprimed  (\tabref{misprimed} C) LAMA.}

   \end{table*}

  \textbf{Misprimed LAMA.} \tabref{misprimed} shows
  the effect of mispriming on BERT-large for
questions answered
  correctly in original LAMA; recall that
  \tabref{misprimed_examples} gives examples of sentences
  constructed in modes A, B, C and D.
  In most cases,
  mispriming with a highly ranked incorrect object causes a precision
  drop of over 60\% (C).  Example predictions can be found
  in \tabref{examples} (lines marked ``M''). This
  sensitivity to misprimes still exists when  the distance
  between misprime and   cloze
  question is increased: the drop persists when
  20 sentences are inserted (D).
  Striking are the results for  Google-RE  where
  the model recalls almost no facts (C).
  \tabref{examples} (lines marked ``M'') shows predicted fillers for these
  misprimed sentences.  BERT is less but still badly
  affected by misprimes  that 
match selectional  restrictions  (B).  The model is
  more robust against priming with random words
  (A):
the precision drop 
is on average more than 35\% lower than for (D).
  We included the baseline (A) as a sanity check
  for the precision drop measure.
  These baseline results show that the presence of a
  misprime per se does not confuse the model;  a
   less distracting misprime (different type of entity or a
completely  implausible  answer)
often results in a correct answer by BERT.

 \begin{table}
  \small
  \centering
\begin{tabular}{l|ll|ll}
&\multicolumn{2}{c}{\textbf{train}} & \multicolumn{2}{c}{\textbf{test}}\\
& pos& neg&pos & neg  \\ \hline
pretrained BERT & 0.9& 0.9& 0.2&0.2   \\
finetuned BERT & 1.0& 1.0&1.0 &1.0
\end{tabular}
\caption{\tablabel{retrain} Accuracy of BERT on balanced corpus.
Pretrained BERT does not model negation well, but finetuned
BERT  classifies sentences as true/false correctly.}
\end{table}

\enote{hs}{add arugment: psychological priming is also artificial}

 \section{Discussion}
Whereas
\citet{petroni2019language}'s results
suggest that PLMs are able to memorize
facts,
our results indicate that PLMs largely do not learn the
meaning of negation.
They mostly seem to predict
fillers based on co-occurrence of subject (e.g., ``Quran'')
and filler (``religious'')
and to ignore negation.

A key problem is that in the LAMA setup, not answering
(i.e., admitting ignorance) is not an option.
While the prediction
probability generally is somewhat lower in the negated
compared to the positive answer, there is no threshold
across cloze questions that could be used to distinguish
valid positive from invalid negative answers (cf.\ \tabref{examples}).

We suspect that
a possible explanation for PLMs' poor performance is
that negated sentences occur much less frequently in training corpora.
Our synthetic corpus study
(\tabref{retrain})
shows that BERT is able to memorize
negative facts that occur in the corpus.
However, the
PLM objective encourages the
model
to predict fillers based  on similar sentences in the
training corpus -- and if the most similar statement to a
negative sentence is positive, then the filler is generally incorrect.
However,
after finetuning, BERT is able to classify
truth/falseness correctly, demonstrating that negation can be
learned through supervised training.

The mispriming experiment shows that BERT often  handles
random misprimes correctly (\tabref{misprimed} A).
There are also cases where BERT does the right thing for
difficult misprimes, e.g.,
it robustly
attributes ``religious'' to Quran (\tabref{examples}).
In general, however,
BERT is highly
sensitive to misleading context (\tabref{misprimed} C) that
would not  change human behavior in QA.
It is especially striking that a single word suffices to
distract BERT.
This may suggest that
it is not knowledge that is learned by BERT, but that its
performance is mainly based on similarity matching between
the current context  on the one hand and sentences in its
training corpus and/or recent context on the other hand.
\citet{poerner2019bert}
present a similar analysis.

Our work is a new way of analyzing differences
between PLMs
and human-level natural language
understanding. We should aspire to develop PLMs that
-- like humans --
can handle negation and are not easily distracted by misprimes.

\section{Related Work}
PLMs are top performers for many tasks,
including QA \cite{qa,Alberti2019ABB}. PLMs are usually
finetuned \cite{liu2019roberta, devlin-etal-2019-bert}, but
recent work has applied models without  finetuning
\cite{Radford2019LanguageMA,
  petroni2019language}. \citet{DBLP:journals/corr/abs-1906-05317}
investigate
PLMs'
common sense knowledge, but do not consider negation explicitly or priming.

A wide range of literature analyzes linguistic knowledge stored in pretrained embeddings  \cite{jumelet-hupkes-2018-language, gulordava-etal-2018-colorless, giulianelli-etal-2018-hood, mccoy-etal-2019-right, 607823, MarvinLinzen2018, unknown4, kann-etal-2019-verb}. Our work analyzes factual knowledge.
\citet{mccoy-etal-2019-right} show that BERT finetuned to  perform  natural language inference heavily relies on syntactic heuristics, also suggesting that it is not able to adequately acquire common sense.

\citet{unknown2} investigate BERT's understanding of
how negative polarity items are licensed.
Our work,
focusing on factual knowledge stored in negated
sentences, is complementary since grammaticality and
factuality are mostly orthogonal properties. \citet{unknown3} investigate
understanding of negation particles
when PLMs are finetuned.
In contrast, our focus is on
the interaction of negation and factual knowledge learned in pretraining.
\citet{unknown1} defines and applies psycho-linguistic diagnostics
for PLMs. Our use of priming is complementary.
Their data consists of two sets of 72
and 16 sentences whereas we create
42,867 negated sentences covering a wide
range of topics and relations.

\citet{ribeiro-etal-2018-semantically} test for
comprehension of minimally modified sentences in an
adversarial setup while trying to keep the overall semantics
the same.  In contrast, we investigate large changes of
meaning (negation) and context (mispriming).  In contrast to
adversarial work
(e.g., \citep{Wallace2019Triggers}),
we do not focus on adversarial examples for a specific task, but on
pretrained models' ability to
robustly store factual knowledge.

\section{Conclusion}

Our results suggest that
pretrained language models address open domain QA in
datasets like LAMA by mechanisms that are more akin to
relatively shallow pattern matching than the recall of learned factual knowledge
and inference.

\medskip

\textbf{Implications for future work on pretrained language
  models.}
(i) Both factual knowledge and logic are discrete phenomena in
the sense that sentences with similar representations in
current pretrained language models differ sharply in
factuality and truth value (e.g., ``Newton was born in
1641'' vs.\ ``Newton was born in 1642'').  Further
architectural innovations in deep learning seem necessary to
deal with such discrete phenomena.
(ii)
We found that
PLMs have difficulty
distinguishing ``informed'' best guesses (based on
information extracted from training corpora) from ``random''
best guesses (made in the absence of any evidence in the
training corpora). This implies that better confidence assessment
of PLM predictions is needed.
(iii) Our premise was that we should emulate human language
processing and that therefore tasks that are easy for humans
are good tests for NLP models. To the extent this is true,
the two phenomena we have investigated in this paper --
that PLMs seem to ignore negation in many cases and that
they are easily confused by simple distractors 
-- seem to be good vehicles for encouraging the development
of PLMs whose performance on NLP tasks is closer to humans.

\bigskip

\textbf{Acknowledgements.}
We thank the
        reviewers for their constructive criticism.
This work was funded by the German Federal Ministry of Education and
Research (BMBF) under Grant No.\ 01IS18036A
and
        by the European Research Council (Grant No.\ 740516). The authors of this work take full
        responsibility for its content.

\bibliography{anthology,acl2020}

\begin{thebibliography}{28}
\expandafter\ifx\csname natexlab\endcsname\relax\def\natexlab#1{#1}\fi

\bibitem[{Alberti et~al.(2019)Alberti, Lee, and Collins}]{Alberti2019ABB}
Chris Alberti, Kenton Lee, and Michael Collins. 2019.
\newblock A {BERT} baseline for the natural questions.
\newblock \emph{ArXiv}, abs/1901.08634.

\bibitem[{Bosselut et~al.(2019)Bosselut, Rashkin, Sap, Malaviya, Celikyilmaz,
  and Choi}]{DBLP:journals/corr/abs-1906-05317}
Antoine Bosselut, Hannah Rashkin, Maarten Sap, Chaitanya Malaviya, Asli
  Celikyilmaz, and Yejin Choi. 2019.
\newblock \href {https://doi.org/10.18653/v1/P19-1470} {{COMET}: Commonsense
  transformers for automatic knowledge graph construction}.
\newblock In \emph{Proceedings of the 57th Annual Meeting of the Association
  for Computational Linguistics}, pages 4762--4779, Florence, Italy.
  Association for Computational Linguistics.

\bibitem[{Dai et~al.(2019)Dai, Yang, Yang, Carbonell, Le, and
  Salakhutdinov}]{unknown}
Zihang Dai, Zhilin Yang, Yiming Yang, Jaime Carbonell, Quoc Le, and Ruslan
  Salakhutdinov. 2019.
\newblock \href {https://doi.org/10.18653/v1/P19-1285} {Transformer-{XL}:
  Attentive language models beyond a fixed-length context}.
\newblock In \emph{Proceedings of the 57th Annual Meeting of the Association
  for Computational Linguistics}, pages 2978--2988, Florence, Italy.
  Association for Computational Linguistics.

\bibitem[{Dasgupta et~al.(2018)Dasgupta, Guo, Stuhlm{\"u}ller, Gershman, and
  Goodman}]{607823}
Ishita Dasgupta, Demi Guo, Andreas Stuhlm{\"u}ller, Samuel~J Gershman, and
  Noah~D Goodman. 2018.
\newblock Evaluating compositionality in sentence embeddings.
\newblock \emph{arXiv preprint arXiv:1802.04302}.

\bibitem[{Devlin et~al.(2019)Devlin, Chang, Lee, and
  Toutanova}]{devlin-etal-2019-bert}
Jacob Devlin, Ming-Wei Chang, Kenton Lee, and Kristina Toutanova. 2019.
\newblock \href {https://doi.org/10.18653/v1/N19-1423} {{BERT}: Pre-training of
  deep bidirectional transformers for language understanding}.
\newblock In \emph{Proceedings of the 2019 Conference of the North {A}merican
  Chapter of the Association for Computational Linguistics: Human Language
  Technologies, Volume 1 (Long and Short Papers)}, pages 4171--4186,
  Minneapolis, Minnesota. Association for Computational Linguistics.

\bibitem[{Elsahar et~al.(2018)Elsahar, Vougiouklis, Remaci, Gravier, Hare,
  Laforest, and Simperl}]{DBLP:conf/lrec/ElSaharVRGHLS18}
Hady Elsahar, Pavlos Vougiouklis, Arslen Remaci, Christophe Gravier, Jonathon
  Hare, Frederique Laforest, and Elena Simperl. 2018.
\newblock \href {https://www.aclweb.org/anthology/L18-1544} {T-{RE}x: A large
  scale alignment of natural language with knowledge base triples}.
\newblock In \emph{Proceedings of the Eleventh International Conference on
  Language Resources and Evaluation ({LREC} 2018)}, Miyazaki, Japan. European
  Language Resources Association (ELRA).

\bibitem[{Ettinger(2019)}]{unknown1}
Allyson Ettinger. 2019.
\newblock What {BERT} is not: Lessons from a new suite of psycholinguistic
  diagnostics for language models.
\newblock \emph{Transactions of the Association for Computational Linguistics},
  8:34--48.

\bibitem[{Giulianelli et~al.(2018)Giulianelli, Harding, Mohnert, Hupkes, and
  Zuidema}]{giulianelli-etal-2018-hood}
Mario Giulianelli, Jack Harding, Florian Mohnert, Dieuwke Hupkes, and Willem
  Zuidema. 2018.
\newblock \href {https://doi.org/10.18653/v1/W18-5426} {Under the hood: Using
  diagnostic classifiers to investigate and improve how language models track
  agreement information}.
\newblock In \emph{Proceedings of the 2018 {EMNLP} Workshop {B}lackbox{NLP}:
  Analyzing and Interpreting Neural Networks for {NLP}}, pages 240--248,
  Brussels, Belgium. Association for Computational Linguistics.

\bibitem[{Gulordava et~al.(2018)Gulordava, Bojanowski, Grave, Linzen, and
  Baroni}]{gulordava-etal-2018-colorless}
Kristina Gulordava, Piotr Bojanowski, Edouard Grave, Tal Linzen, and Marco
  Baroni. 2018.
\newblock \href {https://doi.org/10.18653/v1/N18-1108} {Colorless green
  recurrent networks dream hierarchically}.
\newblock In \emph{Proceedings of the 2018 Conference of the North {A}merican
  Chapter of the Association for Computational Linguistics: Human Language
  Technologies, Volume 1 (Long Papers)}, pages 1195--1205, New Orleans,
  Louisiana. Association for Computational Linguistics.

\bibitem[{Jumelet and Hupkes(2018)}]{jumelet-hupkes-2018-language}
Jaap Jumelet and Dieuwke Hupkes. 2018.
\newblock \href {https://doi.org/10.18653/v1/W18-5424} {Do language models
  understand anything? on the ability of {LSTM}s to understand negative
  polarity items}.
\newblock In \emph{Proceedings of the 2018 {EMNLP} Workshop {B}lackbox{NLP}:
  Analyzing and Interpreting Neural Networks for {NLP}}, pages 222--231,
  Brussels, Belgium. Association for Computational Linguistics.

\bibitem[{Kann et~al.(2019)Kann, Warstadt, Williams, and
  Bowman}]{kann-etal-2019-verb}
Katharina Kann, Alex Warstadt, Adina Williams, and Samuel~R. Bowman. 2019.
\newblock \href {https://doi.org/10.7275/q5js-4y86} {Verb argument structure
  alternations in word and sentence embeddings}.
\newblock In \emph{Proceedings of the Society for Computation in Linguistics
  ({SC}i{L}) 2019}, pages 287--297.

\bibitem[{Kim et~al.(2019)Kim, Patel, Poliak, Xia, Wang, McCoy, Tenney, Ross,
  Linzen, Van~Durme, Bowman, and Pavlick}]{unknown3}
Najoung Kim, Roma Patel, Adam Poliak, Patrick Xia, Alex Wang, Tom McCoy, Ian
  Tenney, Alexis Ross, Tal Linzen, Benjamin Van~Durme, Samuel~R. Bowman, and
  Ellie Pavlick. 2019.
\newblock \href {https://doi.org/10.18653/v1/S19-1026} {Probing what different
  {NLP} tasks teach machines about function word comprehension}.
\newblock In \emph{Proceedings of the Eighth Joint Conference on Lexical and
  Computational Semantics (*{SEM} 2019)}, pages 235--249, Minneapolis,
  Minnesota. Association for Computational Linguistics.

\bibitem[{Kwiatkowski et~al.(2019)Kwiatkowski, Palomaki, Redfield, Collins,
  Parikh, Alberti, Epstein, Polosukhin, Devlin, Lee, Toutanova, Jones, Kelcey,
  Chang, Dai, Uszkoreit, Le, and Petrov}]{qa}
Tom Kwiatkowski, Jennimaria Palomaki, Olivia Redfield, Michael Collins, Ankur
  Parikh, Chris Alberti, Danielle Epstein, Illia Polosukhin, Jacob Devlin,
  Kenton Lee, Kristina Toutanova, Llion Jones, Matthew Kelcey, Ming-Wei Chang,
  Andrew~M. Dai, Jakob Uszkoreit, Quoc Le, and Slav Petrov. 2019.
\newblock \href {https://doi.org/10.1162/tacl_a_00276} {Natural questions: A
  benchmark for question answering research}.
\newblock \emph{Transactions of the Association for Computational Linguistics},
  7:453--466.

\bibitem[{Li et~al.(2016)Li, Taheri, Tu, and Gimpel}]{li-etal-2016-commonsense}
Xiang Li, Aynaz Taheri, Lifu Tu, and Kevin Gimpel. 2016.
\newblock \href {https://doi.org/10.18653/v1/P16-1137} {Commonsense knowledge
  base completion}.
\newblock In \emph{Proceedings of the 54th Annual Meeting of the Association
  for Computational Linguistics (Volume 1: Long Papers)}, pages 1445--1455,
  Berlin, Germany. Association for Computational Linguistics.

\bibitem[{Liu et~al.(2019)Liu, Ott, Goyal, Du, Joshi, Chen, Levy, Lewis,
  Zettlemoyer, and Stoyanov}]{liu2019roberta}
Yinhan Liu, Myle Ott, Naman Goyal, Jingfei Du, Mandar Joshi, Danqi Chen, Omer
  Levy, Mike Lewis, Luke Zettlemoyer, and Veselin Stoyanov. 2019.
\newblock {RoBERTa}: {A} robustly optimized {BERT} pretraining approach.
\newblock \emph{arXiv preprint arXiv:1907.11692}.

\bibitem[{Marvin and Linzen(2018)}]{MarvinLinzen2018}
Rebecca Marvin and Tal Linzen. 2018.
\newblock \href {https://doi.org/10.18653/v1/D18-1151} {Targeted syntactic
  evaluation of language models}.
\newblock In \emph{Proceedings of the 2018 Conference on Empirical Methods in
  Natural Language Processing}, pages 1192--1202, Brussels, Belgium.
  Association for Computational Linguistics.

\bibitem[{McCoy et~al.(2019)McCoy, Pavlick, and Linzen}]{mccoy-etal-2019-right}
Tom McCoy, Ellie Pavlick, and Tal Linzen. 2019.
\newblock \href {https://doi.org/10.18653/v1/P19-1334} {Right for the wrong
  reasons: Diagnosing syntactic heuristics in natural language inference}.
\newblock In \emph{Proceedings of the 57th Annual Meeting of the Association
  for Computational Linguistics}, pages 3428--3448, Florence, Italy.
  Association for Computational Linguistics.

\bibitem[{Peters et~al.(2018)Peters, Neumann, Iyyer, Gardner, Clark, Lee, and
  Zettlemoyer}]{peters-etal-2018-deep}
Matthew Peters, Mark Neumann, Mohit Iyyer, Matt Gardner, Christopher Clark,
  Kenton Lee, and Luke Zettlemoyer. 2018.
\newblock \href {https://doi.org/10.18653/v1/N18-1202} {Deep contextualized
  word representations}.
\newblock In \emph{Proceedings of the 2018 Conference of the North {A}merican
  Chapter of the Association for Computational Linguistics: Human Language
  Technologies, Volume 1 (Long Papers)}, pages 2227--2237, New Orleans,
  Louisiana. Association for Computational Linguistics.

\bibitem[{Petroni et~al.(2019)Petroni, Rockt{\"a}schel, Riedel, Lewis, Bakhtin,
  Wu, and Miller}]{petroni2019language}
Fabio Petroni, Tim Rockt{\"a}schel, Sebastian Riedel, Patrick Lewis, Anton
  Bakhtin, Yuxiang Wu, and Alexander Miller. 2019.
\newblock \href {https://doi.org/10.18653/v1/D19-1250} {Language models as
  knowledge bases?}
\newblock In \emph{Proceedings of the 2019 Conference on Empirical Methods in
  Natural Language Processing and the 9th International Joint Conference on
  Natural Language Processing (EMNLP-IJCNLP)}, pages 2463--2473, Hong Kong,
  China. Association for Computational Linguistics.

\bibitem[{Poerner et~al.(2019)Poerner, Waltinger, and
  Sch{\"u}tze}]{poerner2019bert}
Nina Poerner, Ulli Waltinger, and Hinrich Sch{\"u}tze. 2019.
\newblock {BERT} is not a knowledge base (yet): Factual knowledge vs.
  name-based reasoning in unsupervised qa.
\newblock \emph{ArXiv}, abs/1911.03681.

\bibitem[{Radford et~al.(2019)Radford, Wu, Child, Luan, Amodei, and
  Sutskever}]{Radford2019LanguageMA}
Alec Radford, Jeffrey Wu, Rewon Child, David Luan, Dario Amodei, and Ilya
  Sutskever. 2019.
\newblock Language models are unsupervised multitask learners.

\bibitem[{Rajpurkar et~al.(2016)Rajpurkar, Zhang, Lopyrev, and
  Liang}]{rajpurkar-etal-2016-squad}
Pranav Rajpurkar, Jian Zhang, Konstantin Lopyrev, and Percy Liang. 2016.
\newblock \href {https://doi.org/10.18653/v1/D16-1264} {{SQ}u{AD}: 100,000+
  questions for machine comprehension of text}.
\newblock In \emph{Proceedings of the 2016 Conference on Empirical Methods in
  Natural Language Processing}, pages 2383--2392, Austin, Texas. Association
  for Computational Linguistics.

\bibitem[{Ribeiro et~al.(2018)Ribeiro, Singh, and
  Guestrin}]{ribeiro-etal-2018-semantically}
Marco~Tulio Ribeiro, Sameer Singh, and Carlos Guestrin. 2018.
\newblock \href {https://doi.org/10.18653/v1/P18-1079} {Semantically equivalent
  adversarial rules for debugging {NLP} models}.
\newblock In \emph{Proceedings of the 56th Annual Meeting of the Association
  for Computational Linguistics (Volume 1: Long Papers)}, pages 856--865,
  Melbourne, Australia. Association for Computational Linguistics.

\bibitem[{Tulving and Schacter(1990)}]{Tulving301}
Endel Tulving and Daniel Schacter. 1990.
\newblock \href {https://doi.org/10.1126/science.2296719} {Priming and human
  memory systems}.
\newblock \emph{Science}, 247(4940):301--306.

\bibitem[{Wallace et~al.(2019)Wallace, Feng, Kandpal, Gardner, and
  Singh}]{Wallace2019Triggers}
Eric Wallace, Shi Feng, Nikhil Kandpal, Matt Gardner, and Sameer Singh. 2019.
\newblock \href {https://doi.org/10.18653/v1/D19-1221} {Universal adversarial
  triggers for attacking and analyzing {NLP}}.
\newblock In \emph{Proceedings of the 2019 Conference on Empirical Methods in
  Natural Language Processing and the 9th International Joint Conference on
  Natural Language Processing (EMNLP-IJCNLP)}, pages 2153--2162, Hong Kong,
  China. Association for Computational Linguistics.

\bibitem[{Warstadt and Bowman(2019)}]{unknown4}
Alex Warstadt and Samuel~R. Bowman. 2019.
\newblock Grammatical analysis of pretrained sentence encoders with
  acceptability judgments.
\newblock \emph{ArXiv}, abs/1901.03438.

\bibitem[{Warstadt et~al.(2019)Warstadt, Cao, Grosu, Peng, Blix, Nie, Alsop,
  Bordia, Liu, Parrish, Wang, Phang, Mohananey, Htut, Jeretic, and
  Bowman}]{unknown2}
Alex Warstadt, Yu~Cao, Ioana Grosu, Wei Peng, Hagen Blix, Yining Nie, Anna
  Alsop, Shikha Bordia, Haokun Liu, Alicia Parrish, Sheng-Fu Wang, Jason Phang,
  Anhad Mohananey, Phu~Mon Htut, Paloma Jeretic, and Samuel~R. Bowman. 2019.
\newblock \href {https://doi.org/10.18653/v1/D19-1286} {Investigating
  {BERT}{'}s knowledge of language: Five analysis methods with {NPI}s}.
\newblock In \emph{Proceedings of the 2019 Conference on Empirical Methods in
  Natural Language Processing and the 9th International Joint Conference on
  Natural Language Processing (EMNLP-IJCNLP)}, pages 2877--2887, Hong Kong,
  China. Association for Computational Linguistics.

\bibitem[{Wolf et~al.(2019)Wolf, Debut, Sanh, Chaumond, Delangue, Moi, Cistac,
  Rault, Louf, Funtowicz, and Brew}]{Wolf2019HuggingFacesTS}
Thomas Wolf, Lysandre Debut, Victor Sanh, Julien Chaumond, Clement Delangue,
  Anthony Moi, Pierric Cistac, Tim Rault, R'emi Louf, Morgan Funtowicz, and
  Jamie Brew. 2019.
\newblock Huggingface's transformers: State-of-the-art natural language
  processing.
\newblock \emph{ArXiv}, abs/1910.03771.

\end{thebibliography}
\bibliographystyle{acl_natbib}

\section{Appendix}
\subsection{Details on LAMA}
We use source code provided by \citet{petroni2019language} \footnote{\url{github.com/facebookresearch/LAMA}}.
T-REx, parts of ConceptNet and SQuAD allow multiple true answers (N-M).
To ensure single true
objects for Google-RE, we reformulate the templates asking for location to
specifically ask for cities (e.g., ``born in [MASK]'' to ``born
in the city of [MASK]''). We do not change any other templates. T-REx still queries for "died in [MASK]".
\subsubsection{Details on negated LAMA}
For ConceptNet we extract an easy-to-negate subset. 
The final subset includes 2,996 of the 11,458 samples.
We proceed as follows:

1. We only negate sentences of maximal token sequence length 4 or if we find a match with one of the following patterns:
``is a type of '', ``is made of'', ``is part of'', 
``are made of.'', ``can be made of'', ``are a type of '', ``are a part off''.

2. The selected subset is automatically negated by a manually created verb negation dictionary.

\subsubsection{Details on misprimed LAMA}
To investigate the effect of distance between the prime and the cloze question, we insert a concatenation of up to 20 ``neutral'' sentences. The longest sequence has 89 byte pair encodings. The distance upon the full concatenation of all 20 sentences did not lessen the effect of the prime much. The used sentences are:
"This is great.", "This is interesting.", "Hold this thought.",
 "What a surprise.", "Good to know.", "Pretty awesome stuff.",
 "Nice seeing you.", "Let's meet again soon.", "This is nice.",
 "Have a nice time.", "That is okay.", "Long time no see.",
 "What a day.", "Wonderful story.", "That's new to me.", "Very cool.",
 "Till next time.", "That's enough.", "This is amazing.",
 "I will think about it."

\begin{table}[h!]
\begin{tabular}{lll}
batch size & 512\\\hline
learning rate & 6e-5\\\hline
number of epochs& 2000\\\hline
max. sequence length& 13
\end{tabular}
\caption{\tablabel{pretraining}Hyper-parameters for pretraining BERT-base on a balanced corpus of negative and positive sentences.}
\end{table}
\begin{table}[h!]
\begin{tabular}{lll}
batch size & 32\\\hline
learning rate & 4e-5\\\hline
number of epochs& 20\\\hline
max. sequence length& 7
\end{tabular}
\caption{\tablabel{finetune}Hyper-parameters for finetuning on the task of classifying
sentences as true/false.}
\end{table}

\begin{figure}
  \includegraphics[width=\linewidth]{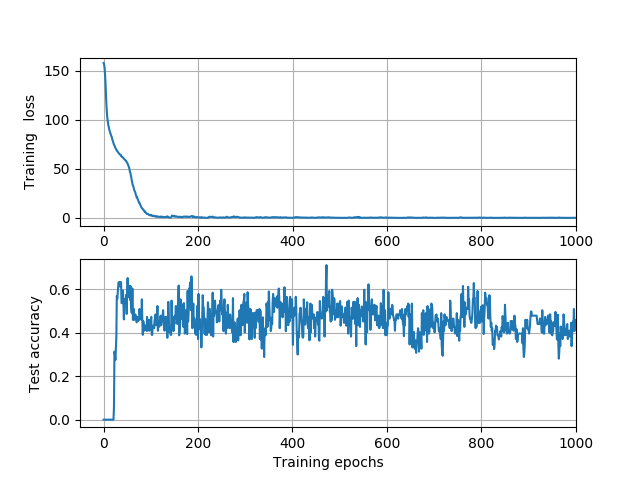}
  \caption{ \figlabel{loss}Training loss and test accuracy when pretraining BERT-base on a balanced corpus. The model is able to memorize positive and negative sentences seen during training but is not able to generalize to an unseen test set for both positive and negative sentences. }
 
\end{figure}

\subsection{Details on the balanced corpus}
We pretrain BERT-base from scratch on a corpus on equally many negative and positive sentences. 
We concatenate multiples of the same training data into one training file to compensate for the little amount of data. Hyper-parameters for pretraining are listed in \tabref{pretraining}. The full vocabulary is 349 tokens long. 

\figref{loss} shows that training loss and test accuracy are uncorrelated. Test accuracy stagnates around 0.5 which is not more than random guessing as for each relation half of the adjectives hold.

We finetune
on the task of classifying
sentences as true/false. We concatenate multiples of the same training data into one training file to compensate for the little amount of data. 
Hyperparameters for finetuning are listed in \tabref{finetune}.

We use source code provided by \citet{Wolf2019HuggingFacesTS} \footnote{\url{github.com/huggingface/transformers}}.

\end{document}